# Early Detection of At-Risk Students Using Machine Learning

Azucena L. Jimenez Martinez*,§, Kanika Sood*,‡, and Rakeshkumar Mahto†,‡

*Department of Computer Science
†Department of Electrical and Computer Engineering
California State University
Fullerton, California, USA
‡{kasood, ramahto}@fullerton.edu
§azucenaljim1@csu.fullerton.edu

*Abstract*—This research presents preliminary work to address the challenge of identifying at-risk students using supervised machine learning and three unique data categories: engagement, demographics, and performance data collected from Fall 2023 using Canvas and the California State University, Fullerton dashboard. We aim to tackle the persistent challenges of higher education retention and student dropout rates by screening for at-risk students and building a high-risk identification system. By focusing on previously overlooked behavioral factors alongside traditional metrics, this work aims to address educational gaps, enhance student outcomes, and significantly boost student success across disciplines at the University. Pre-processing steps take place to establish a target variable, anonymize student information, manage missing data, and identify the most significant features. Given the mixed data types in the datasets and the binary classification nature of this study, this work considers several machine learning models, including Support Vector Machines (SVM), Naive Bayes, K-nearest neighbors (KNN), Decision Trees, Logistic Regression, and Random Forest. These models predict at-risk students and identify critical periods of the semester when student performance is most vulnerable. We will use validation techniques such as train test split and k-fold cross-validation to ensure the reliability of the models. Our analysis indicates that all algorithms generate an acceptable outcome for at-risk student predictions, while Naive Bayes performs best overall.

## I. Introduction

Student dropout rates are a significant concern in education and policy circles. This research seeks to highlight the complexities surrounding identifying at-risk students and develop a machine learning model to screen for these students, providing instructors with a tool to intervene early and ensuring equitable educational opportunities for all students. It reviews the implementation and advancement of artificial intelligence and machine learning in developing a prediction model for at-risk students. It illustrates the necessity for instructors to have a reliable way to intervene and provide students with academic assistance. Over the past few years, several related studies have focused on using artificial intelligence and machine learning techniques to create risk prediction systems for students at a higher education level. Past research has shown that a dropout early warning system enables schools to identify students who are at risk preemptively. This project will contribute to academic success by providing timely interventions, individualized assistance, an environment for student feedback, and targeted support to overcome challenges and achieve educational goals.

Traditional methods of identifying at-risk students, such as focusing on grades and attendance records, may need to be revised to capture the complex factors contributing to student success. The key to understanding the factors that may influence dropout behavior is recognizing that dropping out is a process rather than an isolated event [1]. Many attributes, such as schooling, individual personality traits, home environment, and economic context, all influence a student's progress [1]. Each year, roughly 30% of first-year students at US baccalaureate institutions do not return for their second year, and over $9 billion is spent educating these students [1]. Examining individual features reveals several interesting trends. Firstly, GPA in math, English, chemistry, and psychology classes were among the most vital individual predictors of student retention. California State University, Fullerton's (CSUF) statistics on graduating students are fairly average. CSUF offers 57 bachelor's degrees with an average graduation rate of 69%. The university's average first to second-year retention rate is 70.57% for four years or less and 62.7% for five years or more. Based on previous studies mentioned under related work, many predictive and risk models have been built to identify students who are at risk; however, a few still lack focus on behavioral and background factors in the development of such predictive models.

This study is conducted with participating Engineering and Computer Science faculty. All participating members obtain institutional Review Board (IRB) certification and actively participate in the data collection and training workshops in the academic year 2023-2024. Their involvement ensures a comprehensive and ethical approach to gathering and analyzing the data, enhancing the study's credibility and relevance. This collaborative effort allows for a real-time understanding of student performance and the effectiveness of the predictive model in identifying at-risk students.

## II. BACKGROUND WORK

Over the past few years, several related studies have focused on using artificial intelligence and machine learning techniques in effective risk prediction systems for students at a higher education level. While previous studies overlook behavioral factors, our study incorporates three unique data categories: engagement, demographics, and grade data. These factors are crucial in addressing educational gaps and improving student performance and graduation rates.

This research contributes to the CSU system's Graduation Initiative for 2025-2030, which seeks to boost graduation rates and reduce achievement gaps [2]. It will contribute to academic success by providing timely interventions, individualized assistance, an environment for student feedback, and targeted support to overcome challenges and achieve educational goals. Identifying and predicting a student at risk of failing or dropping out may be challenging for an instructor who keeps track of multiple students on multiple days. It is not possible for an instructor to offer personalized assistance without the student reaching out first. Due to this issue, struggling students who do not reach out first may be disadvantaged regarding their education rates.

### A. Identify At-Risk Students in an Introductory Programming Course at a Two-year Public College

The study aims to improve student success rates using supervised machine learning to identify students who are "at risk" of not succeeding in CS1 (an introductory course) at a two-year public college. It identifies academic-based early alert triggers for CS1, reasoning that the first two graded programs are paramount for student success in the course. The outcome of this pilot study was a 23% increase in student success and a 7.3% drop in the DFW rate (i.e., the percentage of students who receive a D, receive an F, or withdraw) [3]. It identifies the best-performing neural network on an out-of-sample dataset as a Probabilistic neural network (PNN). Using a threshold value of 0.51 for the PNN output results in a sizable increase in predictive accuracy to 99.2% [3]. This study applies data collection of CS1 student records from the instructor's grade books to create the training and testing dataset for the predictive system [3].

### B. The Machine Learning-Based Dropout Early Warning System for Improving the Performance of Dropout Prediction

The present study aims to improve the performance of a dropout early warning system by addressing the class imbalance issue using the synthetic minority oversampling techniques (SMOTE). An early intervention informed by the dropout early warning system can redirect potential dropout students onto the path to graduation and lead them to a better future. The target label for prediction is students' dropouts. The prediction model found that dropout students are more likely to be problematic in attendance and achievement and are less likely to participate in school activities [4]. They report the AUC of the ROC curves for the four binary classifiers Random Forest (RF), boosted decision tree (BDT), RF with SMOTE (SMOTE + RF), and boosted decision tree with SMOTE (SMOTE + BDT) were 0.986, 0.988, 0.986, and 0.991 [4]. The AUC of the PR curves for the four binary classifier of the RF, boosted decision tree (BDT), RF with SMOTE (SMOTE + RF), and boosted decision tree with SMOTE (SMOTE + BDT) were 0.634, 0.898, 0.643, and 0.724. This study uses data samples of 165,715 high school students from the NEIS database of 2014 from two major cities and two provinces in South Korea: Seoul, Incheon, Gyeongsangbuk-do, and Gyeongsangnam-do [4].

### C. An Early Warning System to Detect At-risk Students in Online Higher Education

The study provides an in-depth analysis of a predictive model to detect at-risk students. A method to determine a threshold for evaluating the quality of the predictive model is established. Lastly, an early warning system is developed and tested in a real educational setting being accurate and useful for its purpose to detect at-risk students in online higher education [5]. The prediction outcome for the submodels is to fail the course. This binary variable has two possible values: pass or fail based on the last graded assessment activities. Outcomes report that in the TPR case, the LOESS regression compared to the NB improves from 44.60–86.04% to 47.88–88.43%. However, considerable improvement is obtained with the second algorithm, where the LOESS regression increases until 58.79–93.60% [5]. The naive bayes classification algorithm is the best algorithm to be used in the institution based on the performance observed on the four metrics. This study utilizes data from the database UOC data mart, which stores academic data from the years 2016-2017 [5].

## III. DATASET AND DATA PREPOSSESSING

Data collection is essential for building and training predictive models. All faculty and team members must obtain IRB clearance to submit and use this data. Additionally, students sign a consent form authorizing the use of their data. The collected data is in a .csv or .xlsx file and includes demographic information such as generation status and admission status, performance metrics like assignment scores and GPA, and engagement details such as participation levels and interactions with Canvas. We utilize anonymization measures to ensure privacy. Encryption techniques are needed to anonymize personally identifiable information such as ID (CWID) numbers. Any direct identifiers, such as emails and student names, are removed from the dataset before analysis.

The data this research utilizes is collected for Fall 2023 and Spring 2024 using Canvas and the California State University, Fullerton (CSUF) dashboard. The dataset contains 31 variables with 119 students. Out of those, 21 students were identified as at risk, and 98 students were not identified as at risk. Students are considered at risk if they have a current letter grade of 'D' or lower and have repeated the course. Using these classifications, we create the variable "at-risk." The type variable identifies whether the student screened is at-risk or not. This research uses the column "at-risk" as the target variable for the predictive models. Data collection is ongoing throughout the Spring semester of 2024 in a time series of data analysis from more faculty members. Participating faculty members are asked to incorporate formative assessments into their classes. The spring semester consists of 16 weeks, from January 20th - May 10th. We split the semester into three phases: phase 1 focused on weeks 1-8, phase two on weeks 8-12, and phase Three, weeks 12-16.

## A. Feature Selection

Identifying the most relevant features in a dataset is vital for classifier performance. This research selects the most significant features as predictor variables according to their contribution pertinent to the target variable. The dataset is split into two different sets. Out of the 31 initial features considered in the dataset, the first feature set comprises 25 features. The second resulting feature set comprised the top 10 features correlating to the target variable. The techniques investigated are a mixture of filter and wrapper methods. The filter method, correlation, is selected due to its inexpensive computation. The wrapper method, Random Forest, was chosen due to its accuracy performance in over-filtering techniques. In this research, we consider and utilize the top 10 features in the dataset, shown in Table 1, to train with the chosen techniques.

## IV. METHODOLOGY

The prediction model for this research aims to provide reliable predictions for at-risk students by following the flow chart in Figure 1. We compare the different single and ensemble classification models in exhaustive experiments to identify the most effective model for classifying at-risk and not-at-risk students. We examine model performance with and without sampling techniques. The experimentation methodology can be summarized by the following steps:

1) Data preprocessing including the following:
   - Creating the variable "at-risk" and selecting it as our target variable.
   - Handle missing values using single imputation.
   - Apply feature selection techniques to obtain only the most relevant features
2) Model validation using the train test split procedure to simulate the model's performance with new data along with the k-fold cross-validation procedure as a standard method for estimating the machine learning algorithm's performance on a dataset with a 10-times k-fold. The selected single and ensemble classification models are applied using the selected features as the independent variables and the prediction ( at-risk, not at-risk) as the output variable.

## A. Machine learning Modeling

Given the mixed data types in the datasets and the binary classification nature of this study, we consider several machine learning models, such as Support Vector Machines (SVM), Naive Bayes (NB), K-nearest neighbors (KNN), Logistic Regression (LR), Decision Trees (DT), and Random Forest (RF).

TABLE I
REDUCED FEATURE SET OF TOP 10 VARIBALES

| Feature Name | Feature Description | Collection Method |
|---|---|---|
| Current Score | The student's current grade or score in the course, typically calculated as a percentage based on the assignments, quizzes, and exams completed so far | Performance Data |
| Assignment Missing | Indicates whether the student has any assignments that have not been submitted by the due date | Demographic Data |
| GPA | The cumulative grade point average of the student across all courses taken. This is a measure of the student's overall academic performance. | Demographic Data |
| Units Earned | The number of academic credits the student has completed and earned towards their degree | Demographic Data |
| Page Views | The number of times the student has accessed course materials or pages on Canvas. This metric is often used to gauge student engagement and activity within the course platform | Engagement Data |
| Participation | The number of interactive activities the student has participated in, such as discussion posts, collaborations, or other course engagements tracked on Canvas. | Engagement Data |
| Program Action | Specific actions or statuses related to the student's academic program, such as admissions, probation, suspension, or other administrative decisions affecting the student's academic progress. | Demographic Data |
| Assignment on Time | Indicates whether the student submits their assignments by the due date and the percentage of assignments the student has submitted on time out of the total assignments assigned. | Engagement Data |
| Student Engagement | A measure of how actively the student is involved in the course. | Demographic Data |
| Units Attempting | The number of academic credits the student is currently enrolled in and attempting to complete in the current term. This can indicate the student's course load and academic commitment. | Demographic Data |

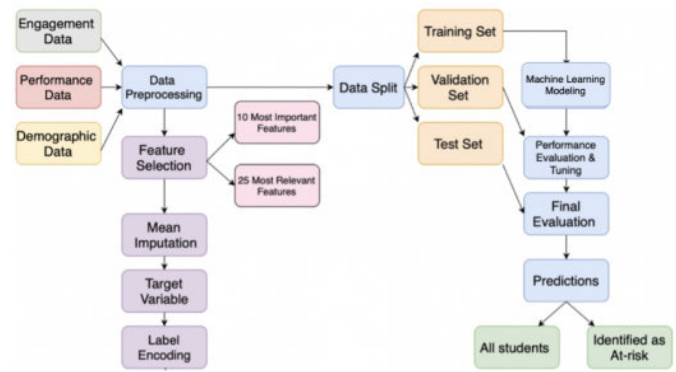

Fig. 1. Machine Learning Flowchart

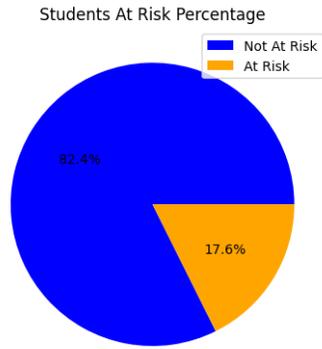

Fig. 2. Data Categories for Student Performance

We first utilize the data without the application of sampling techniques. This means that the instances of at-risk and non-at-risk students are imbalanced, as shown in Figure 2. The predictive outcome of the models considers the imbalanced count of 21 individuals identified as at risk and 98 individuals not identified as at risk with an 80:20 split. To fix the imbalance of data issue, this research uses adaptive synthetic sampling (ADASYN) and synthetic minority (SMOTE) techniques.

*B. Performance Metrics*

Evaluation of classification models is done using the following performance metrics:

- **Overall Accuracy**: The proportion of correct results among all cases.
- **Precision**: The proportion of true positives among predicted positives.
- **Recall**: The proportion of true positives among actual positives.
- **F1-Score**: The harmonic mean of precision and recall.
- **ROC Curve**: A plot of true positive rate vs. false positive rate.

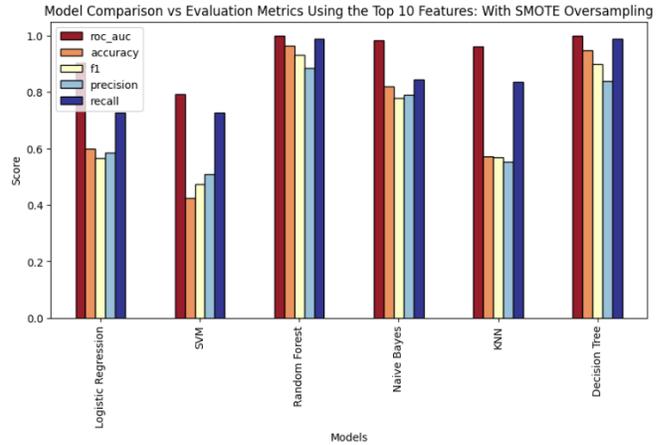

Fig. 4. Model Comparison vs. Evaluation Metrics Using Top 10 Features: SMOTE

Performance metrics in ML are essential for assessing the effectiveness and reliability of models. Figures 3-5 compare the classification models mentioned earlier and their metric performance. These visualizations highlight the strengths and weaknesses of each model, providing a comprehensive evaluation of their ability to predict at-risk students accurately.

## V. RESULTS

In our analysis, we analyze different machine learning and sampling techniques. All ML algorithms generate acceptable prediction outcomes with more than 78% overall accuracy. While all the models perform well, there are slight variations in the prediction accuracy. The top model was Naive Bayes, generating an outcome with more than 89% accuracy.

Tables 2-6 display the classification performance metrics for the previously mentioned classification models. Logistic Regression achieves a mean accuracy of 0.8479, Linear SVM attains 0.8474, KNN reaches 0.8418, Naive Bayes achieves

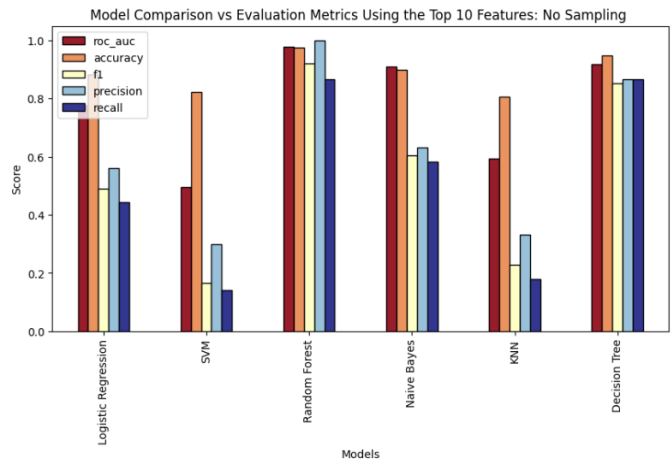

Fig. 3. Model Comparison vs. Evaluation Metrics Using Top 10 Features: No Sampling

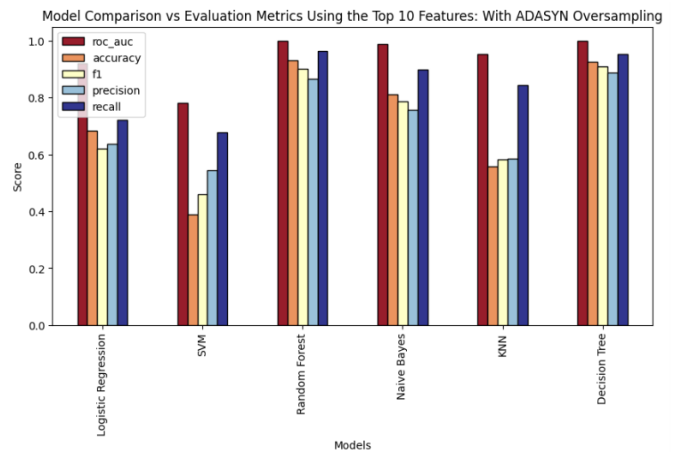

Fig. 5. Model Comparison vs. Evaluation Metrics Using Top 10 Features: ADASYN

TABLE II
CLASSIFICATION PERFORMANCE METRICS FOR LOGISTIC REGRESSION

|  | Precision | Recall | F1-score |
|---|---|---|---|
| not at-risk | 0.88 | 0.89 | 0.88 |
| at-risk | 0.89 | 0.88 | 0.88 |
| Accuracy |  | 0.88 |  |
| Macro Avg | 0.88 | 0.88 | 0.88 |
| Weighted Avg | 0.88 | 0.88 | 0.88 |

TABLE III
CLASSIFICATION PERFORMANCE METRICS FOR LINEAR SVM

|  | Precision | Recall | F1-score |  |
|---|---|---|---|---|
| not at-risk | 0.94 | 0.95 | 0.94 |  |
| at-risk | 0.95 | 0.94 | 0.94 |  |
| Accuracy | 0.94 (196 instances) |  |  |  |
| Macro Avg | 0.94 | 0.94 | 0.94 | 196 |
| Weighted Avg | 0.94 | 0.94 | 0.94 | 196 |

TABLE IV
CLASSIFICATION PERFORMANCE METRICS FOR KNN

|  | Precision | Recall | F1-score |
|---|---|---|---|
| not at-risk | 0.94 | 0.91 | 0.92 |
| at-risk | 0.91 | 0.94 | 0.92 |
| Accuracy | 0.92 (196 instances) |  |  |
| Macro Avg | 0.92 | 0.92 | 0.92 |
| Weighted Avg | 0.92 | 0.92 | 0.92 |

TABLE V
CLASSIFICATION PERFORMANCE METRICS FOR NAIVE BAYES

|  | Precision | Recall | F1-score |
|---|---|---|---|
| not at-risk | 0.89 | 0.95 | 0.92 |
| at-risk | 0.95 | 0.89 | 0.92 |
| Accuracy | 0.92 (196 instances) |  |  |
| Macro Avg | 0.92 | 0.92 | 0.92 |
| Weighted Avg | 0.92 | 0.92 | 0.92 |

0.8937, Decision Tree scores 0.9745, and Random Forest achieves 0.9847 in mean accuracy.

### A. Evaluation of Mispredicted Cases

Despite the small data sample used in this work for building the initial prediction model, consisting of a few sections of classes collected in the fall, is small, the machine learning model we created to predict at-risk students performs well when tested in the Spring with a participating faculty member's classes. Despite the limited data, the model makes accurate predictions, achieving a high success rate with only 10% of mispredictions. The 10% mispredictions statistic is derived from real-time data collected from faculty during different phases of the spring semester. This dataset includes only engagement and demographic data since final grades became available on May 24, after the end of the spring semester. During the semester, faculty received a list of predictions for all students and a separate list identifying only the at-risk students. At the end of phase 3, when grades were released, we received the final grades data, which included information on which students passed the course. We compare the final results to our previous predictions. This comparison revealed that the model accurately identified students' risk status, as evidenced by the 10% misprediction rate. This demonstrates the model's effectiveness and robustness even with a constrained dataset.

### VI. CONCLUSION AND FUTURE WORK

In this work, we apply different sampling techniques to a dataset constructed based on Fall 2023 data collected through Canvas and the CSUF dashboard. Combining different families of ML models to predict the screening of at-risk students. Considering engagement, performance, and demographic factors, this study considers only the features with the highest correlation to the target variable by implementing various feature selection techniques. Our preliminary study showed that all tested ML algorithms could generate acceptable prediction outcomes with more than 78% accuracy. In particular, Random Forest, Naive Bayes, and decision trees achieve overall enhanced performance with more than 89% accuracy.

As future work, this research is ongoing and currently focusing on leveraging Spring 2024 data in a time series data analysis. The Spring 2024 semester is split into three phases: phase 1 focuses on weeks 1-8, phase two on weeks 8-12, and phase three weeks 12-16. Participating faculty members are asked to incorporate formative assessments into their classes. From this, we will analyze whether incorporating formative assessments and the number of at-risk students is correlated. We will also identify the most critical weeks for identifying at-risk students. In the future, we will include creating a tool for instructors to implement and quickly screen for students.

### ACKNOWLEDGMENT

This study is reviewed and approved by the Institutional Review of California State University, Fullerton (Approval Number: HSR-22-23-436), Date: 18th October, 2023. All the students provided written informed consent before using their data for the study. Students who did not give consent were eliminated from the data analysis. This work is supported by California State University's Chancellor's Office through the Creating Responsive, Equitable, Active Teaching and Engagement Awards Program (CREATE) Award.